%% file: main.tex
\documentclass{article}
\PassOptionsToPackage{numbers, compress}{natbib}
\usepackage[final]{neurips_2025}
\usepackage[utf8]{inputenc}
\usepackage[T1]{fontenc}    
\usepackage{hyperref}       
\usepackage{url}            
\usepackage{booktabs}       
\usepackage{amsfonts}       
\usepackage{nicefrac}       
\usepackage{microtype}      
\usepackage{xcolor}         
\usepackage{listings}
\usepackage{enumitem}
\usepackage{mdframed}
\usepackage{comment}
\usepackage{graphicx}
\usepackage{amsmath}
\usepackage{cleveref}
\usepackage{tcolorbox}
\usepackage{lipsum}
\tcbuselibrary{skins}
\usepackage{listings}
\usepackage[htt]{hyphenat}
\usepackage{wrapfig}

\newif\ifarxiv
\arxivtrue    

\ifarxiv
    \usepackage{mathpazo}
    \newcommand{\maybeLooseness}{\looseness=-1}
\else
    \newcommand{\maybeLooseness}{}
\fi

\usepackage[ruled,vlined]{algorithm2e}
\definecolor{customteal}{HTML}{15B38C}
\definecolor{customblue}{HTML}{4D6BFE}
\definecolor{customorange}{HTML}{d97756} 
\definecolor{customgray}{HTML}{777777}

\tcbset{before={\par\vspace*{\baselineskip}},
        after={\par\vspace*{\baselineskip}}}

\definecolor{cornflowerblue}{rgb}{0.39, 0.58, 0.93}
\hypersetup{
    colorlinks=true,
    linkcolor=cornflowerblue,
    filecolor=magenta,      
    urlcolor=teal,
    citecolor=cornflowerblue,
    pdftitle={Antidistillation Sampling},
    pdfauthor={},
    pdfpagemode=FullScreen,
    }

\newcommand{\ntp}[2][x_{t+1}]{p(#1 | x_{1:t}; #2)}
\newcommand{\loss}[1]{\ell\left(#1\right)}

\title{Antidistillation Sampling}
\author{%
Yash Savani\thanks{Equal contribution. Contact: \texttt{\{ysavani,ashert\}@cs.cmu.edu}.}\quad Asher Trockman\footnotemark[1]\quad Zhili Feng\quad Yixuan Even Xu\\ \textbf{Avi Schwarzschild}\quad
\textbf{Alexander Robey}\quad \textbf{Marc Finzi}\quad \textbf{J. Zico Kolter}\\ 
Carnegie Mellon University \\ \\
\url{https://antidistillation.com}
} 

\begin{document}

\maketitle

\begin{abstract}
Frontier models that generate extended reasoning traces inadvertently produce token sequences that can facilitate model distillation. Recognizing this vulnerability, model owners may seek sampling strategies that limit the effectiveness of distillation without compromising model performance. \emph{\bfseries Antidistillation sampling} provides exactly this capability. By strategically modifying a model's next-token probability distribution, antidistillation sampling poisons reasoning traces, rendering them significantly less effective for distillation while preserving the model's utility. Our code is available at \url{https://github.com/locuslab/antidistillation-sampling}.
\end{abstract}

\section{Introduction}
\label{sec:intro}

Large language models (LLMs) trained to produce reasoning traces have achieved strong performance on math, coding, and reasoning benchmarks~\cite{jaech2024openai,claude35sonnet,guo2025deepseek}.  These traces, however, serve a dual purpose.  They not only enhance model performance, but also enable distillation, a process by which a secondary model replicates the original model's capabilities by training on its generations~\cite{hinton2015distilling,schmidhuber1992learning,gou2021knowledge}.  Notably, distillation can result in substantial capability gains at a fraction of the computational cost needed to train similarly performant models from scratch.

While effective and efficient, the viability of distillation poses several downsides for companies deploying frontier reasoning models.  First, returning reasoning traces represents a forfeiture of intellectual property, which can allow competitors to cheaply replicate frontier capabilities.  Second, the threat of distillation incentivizes limiting user access by obscuring token probabilities or truncating reasoning traces.  Finally, model safety is often not preserved by distillation, which enables the generation of harmful content~\cite{sabbaghi2025adversarial,burgess2025deepseek}.

\begin{figure}[t]
    \centering
    \includegraphics[width=\linewidth]{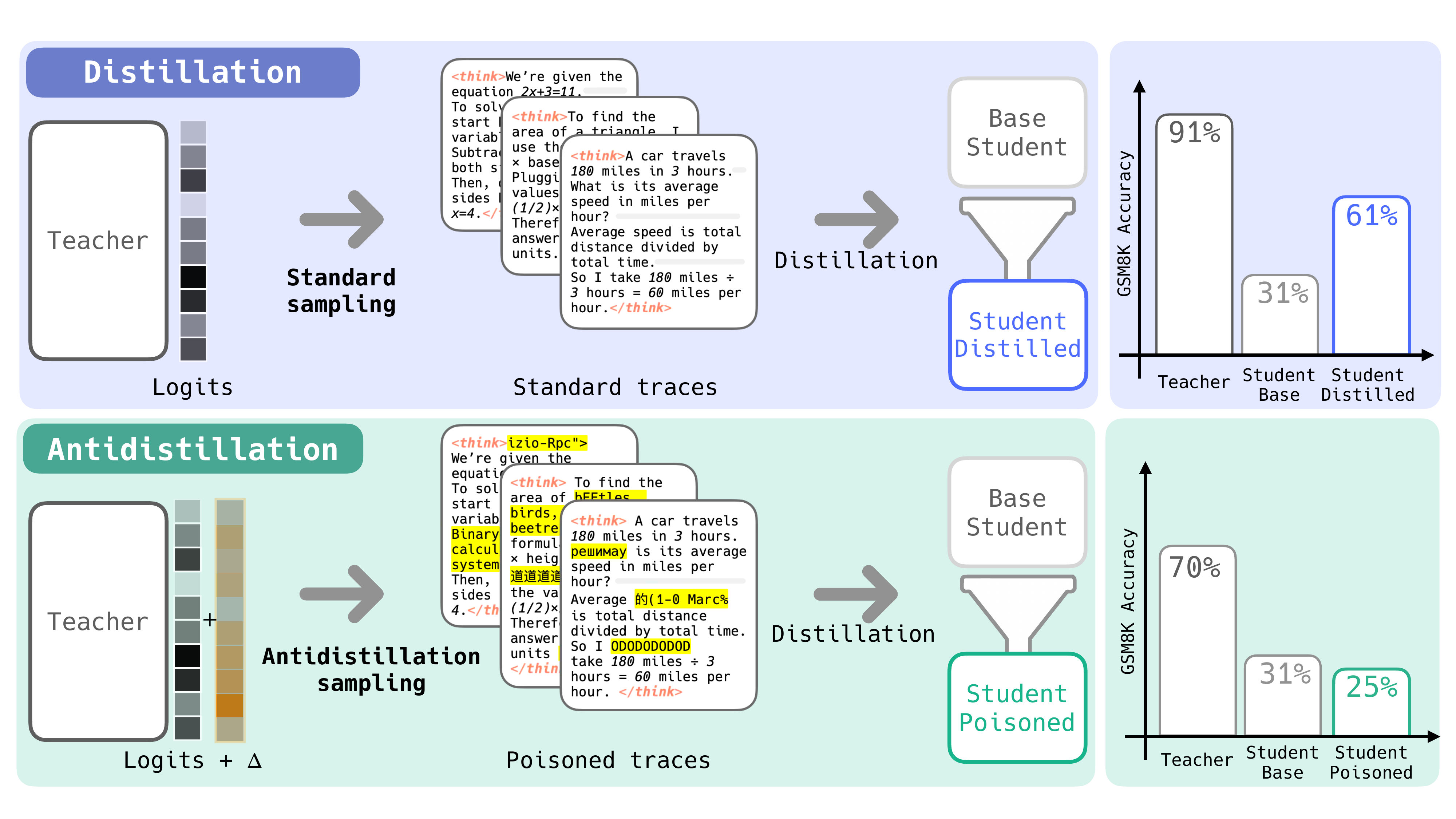}
\caption{Reasoning traces generated via antidistillation sampling poison distillation attempts while simultaneously preserving the teacher's performance. The teacher's logits are perturbed in a direction $\Delta$, leading to samples that significantly degrade distilled model performance relative to naive temperature sampling.  For more details, see \Cref{fig:delta-details} and \S\ref{sec:method}.
}
    \label{fig:banner}
\end{figure}

To address these issues, we introduce \emph{\bfseries antidistillation sampling}~(see \Cref{fig:banner}).  The main idea underpinning antidistillation sampling is to adjust a model's sampling distribution so that generated traces maintain high likelihood under the unadjusted distribution, and distillation attempts are simultaneously poisoned.  To operationalize this idea, we first formulate the general problem of poisoning reasoning models trained via distillation.  We then derive one solution to this problem (see Algorithm~\ref{alg:antidistillation-sampling}), which facilitates a precise trade-off between two competing objectives---the utility of the original model and the effectiveness of distillation poisoning---while incurring minimal computational overhead.

To illustrate our empirical results, consider a reasoning model that achieves 72\% accuracy on MMLU. Naively distilling this model using greedy sampling can produce a student that reaches up to 52\% accuracy. If the teacher model increases its sampling temperature, its accuracy slightly decreases (e.g., by 4\%), yet the distilled student's accuracy remains largely unchanged at 52\%. In contrast, using antidistillation sampling with the same 4\% reduction in teacher accuracy significantly reduces the student's capabilities, lowering its accuracy to as low as 40\%. Our findings on GSM8K~\cite{cobbe2021training}, MATH~\cite{hendrycks2021measuring}, and MMLU~\cite{hendrycks2020measuring} indicate that model owners can effectively limit distillation quality via antidistillation sampling.



\section{Related work}
\label{sec:related-work}

\textbf{Model distillation.} The prominence and effectiveness of model distillation---and, more generally, model compression~\cite{buciluǎ2006model,cheng2017survey,polino2018model}---is rooted in a rich literature dating back to \citet{schmidhuber1992learning}, and, more recently, to~\citet{hinton2015distilling}.  Since these seminal works, a growing body of literature has sought to benchmark the performance of distilled models and to algorithmically maximize the effectiveness of distillation~\cite{shridhar2023distilling, li2024distilling, gu2023minillm, goldblum2020adversarially,romero2014fitnets}.  Indeed, a variety of frontier AI labs have incorporated distillation as a core technique, both to efficiently enable frontier capabilities~\cite{NYTimes2025OpenAIDeepSeek,deepseekai2025deepseekr1incentivizingreasoningcapability} and to improve model safety via context distillation~\cite{bai2022training,snell2022learning,askell2021general}.  However, this practice constitutes a strategic vulnerability for frontier model maintainers, given the demonstrated value of these reasoning traces (see, e.g.,~\cite{jaech2024openai,guan2024deliberative,lightman2023let,uesato2022solving}).

\maybeLooseness
\textbf{Model security.} The setting we address in this paper---where a student model is trained on data generated by a teacher model---intersects with several aspects of model security. For instance, model extraction attacks acquire weights via query-level access, whereas training data extraction attacks are designed to harvest training data~\cite{carlini2024stealing,nasr2023scalable}. While antidistillation sampling may offer some protection against these attacks, such analysis remains beyond our scope. More relevant is the literature on data poisoning, where maliciously crafted data is injected into a model's training set to induce specific downstream effects (see, e.g.,~\cite{goldblum2022dataset}).  In this vein, \citet{tramer2024universal} show the effectiveness of adding backdoors to preference data, sabotaging LLMs finetuned with RLHF.
Our contribution bridges data poisoning and privacy techniques to protect the valuable knowledge encoded in frontier models.

\textbf{Distillation prevention.} A related line of work has sought to develop algorithms that prevent distillation.  In the context of computer vision,~\citet{ma2021undistillable} corrupt the teacher's logits via self-training, whereas follow-up work shows that returning only the top-$k$ logits tends to harm distillation~\cite{ma2022stingy}.  Also related are watermarking algorithms, which seek to adjust model logits to detect whether a model has been distilled~\cite{zhao2023protecting,gu2023learnability,sander2024watermarking,kirchenbauer2023watermark}.  And while this family of methods preserves teacher accuracy, their \emph{static} nature---each input generally yields a deterministic logit vector---presents security vulnerabilities: a distiller can learn an inverse transformation by saving input-output pairs, and thereby fine-tune to recover uncorrupted logits.  To mitigate this shortcoming, \citet{chen2025queen} propose a \emph{session-dynamic} defense that monitors the sensitivity of a user’s queries and perturbs the logits once a threshold is crossed.  In contrast, antidistillation sampling is fully \emph{dynamic}; it perturbs each token’s distribution on-the-fly using gradients from a hidden proxy model, turning generation into a moving target in a similar fashion to cryptographic stream ciphers.

\maybeLooseness
\textbf{Language model decoding.} Finally, we position antidistillation sampling within the broader framework of controlled decoding for LLMs~\cite{mudgal2023controlled}, where supplementary objectives steer the decoding process. Existing approaches in this domain include using contrastive objectives to enhance generation quality~\cite{li2022contrastive}, reformulating constrained decoding as an optimization problem~\cite{ji2023language}, and incorporating energy-based constraints~\cite{qin2022cold}. While related, antidistillation sampling solves a different problem: by implementing a new, distillation-aware penalization term in the decoding objective, our approach poisons generated reasoning traces to undermine the performance of models fine-tuned on these outputs.

\section{Antidistillation sampling}
\label{sec:method}

\begin{algorithm}[t] 
    \label{alg:antidistillation-sampling}
    \DontPrintSemicolon
    \caption{Antidistillation sampling}
    \KwIn{Prompt $x_{1:n}$, max tokens $N$, penalty multiplier $\lambda$, approximation parameter $\epsilon$, temperature $\tau$}

    \vspace{0.5em}

    \qquad 1. (Initialization) Compute the gradient of the downstream loss 
    $$g \gets \nabla\ell(\theta_P)$$

    \vspace{0.5em}
    
    \qquad 2. For each token index $t = n, n+1,\ldots, N-1$:
    \vspace{0.5em}
    
    \qquad \qquad i. Compute the antidistillation penalty term 
    $$\widehat{\Delta}(\:\cdot\:|x_{1:t}) \gets \frac{ \log p(\:\cdot\:|x_{1:t}; \theta_P + \epsilon g) - \log p(\:\cdot\: | x_{1:t}; \theta_P - \epsilon g)}{2\epsilon}$$
    
    \qquad \qquad ii. Sample the next token $x_{t+1}$ from the teacher's adjusted distribution
    $$x_{t+1} \sim \frac{1}{Z}\exp \left( \frac1{\tau} \log p(\:\cdot\:|x_{1:t}; \theta_T) + \lambda \widehat{\Delta}(\:\cdot\:|x_{1:t})\right)$$

    \KwOut{Sampled sequence $x_{1:N}$}
    
\end{algorithm}

To motivate antidistillation sampling, we first sketch a high-level overview of our problem setting in \S\ref{sec:problem-overview}. Based on this setting, we provide a desiderata outlining the desired qualities for poisoning distillation attempts in \S\ref{sec:prelim}. We then derive the antidistillation sampling method (summarized in Algorithm~\ref{alg:antidistillation-sampling}) in \S\ref{sec:antidistillation-sampling-method}.

\subsection{An overview of antidistillation} \label{sec:problem-overview}

The core objective of antidistillation sampling is to adjust a model's next-token distribution to balance two competing goals: sampling tokens with high likelihood under the original, unadjusted distribution and sampling tokens that effectively poison distillation attempts. Throughout, we refer to the model from which reasoning traces are sampled as the \emph{teacher}, and the model being distilled as the \emph{student}.

Our derivation relies on quantifying how model distillation impacts the student model's performance on a given downstream task. This analysis yields a key insight---we can incorporate this performance metric directly into the teacher's sampling distribution. This takes the form of a directional derivative capturing the change in the teacher's sampling distribution along the update direction in the student's weight space. However, due to the high cost needed to compute this directional derivative, the final portion of our derivation identifies an efficient finite-difference approximation for this term, which is inexpensive to compute and, as we demonstrate in \S\ref{sec:results}, results in effective distillation poisoning.

\subsection{Preliminaries}
\label{sec:prelim}

We consider an LLM to be a mapping from a sequence of input tokens $x_{1:t} = (x_1, \dots, x_t)$ to a distribution over the next token, where each token is an element of a vocabulary set $\mathcal{V} = \{1, \dots, V\}$.  This distribution is parameterized by weights $\theta$ and can be expressed as $p(\cdot | x_{1:t}; \theta)$.  We write $p(\cdot | x_{1:t};\theta)$ to denote the distribution of all next-token probabilities, whereas $p(x_{t+1}|x_{1:t};\theta)$ refers to the scalar probability of a given next token $x_{t+1}$. Typically, tokens are generated according to a scaled version of this distribution:\footnote{Variants of this sampling scheme include top-$k$ sampling (i.e., limiting sampling to the tokens with the top-$k$ largest probabilities), greedy sampling (i.e., sampling from the same objective while letting $\tau \rightarrow 0$), and beam search, but we focus mainly on temperature-based sampling here.}
\begin{equation}
x_{t+1} \sim \frac{1}{Z} \exp \left(\frac{1}{\tau} \log p (\cdot | x_{1:t};\theta) \right). \label{eq:sampling}
\end{equation}
Here, $\tau$ is the temperature and $Z$ is a normalization term, which is computed by summing the exponential term over all possible next tokens.  Using a temperature of $\tau=0$ corresponds to greedy sampling, in which $x_{t+1}$ is deterministically chosen to be the token with the largest log probability under the current model parameter $\theta$. 


\textbf{Desiderata for antidistillation.} Distillation involves a student model---parameterized by $\theta_S$, with a distribution over next tokens given by $p(\: \cdot \: |x_{1:t} ; \theta_S)$---trained on data generated from a teacher model parameterized by $\theta_T$. These models do not need to share the same parameter space, and therefore the parameter vectors $\theta_S$ and $\theta_T$ need not be comparable; indeed, a student model may have substantially fewer parameters than the teacher. 
 
The aim of antidistillation sampling is to generate tokens that perform well according to a metric used to the evaluate teacher, while simultaneously having the property that training on these tokens \emph{cannot} improve performance on this same task.
In more detail, we aim to adjust the teacher's sampling procedure to simultaneously satisfy the following:
\begin{enumerate}[label=\Roman*.]
    \item \textbf{Non-distillablity.} Student models trained on tokens sampled via antidistillation sampling should have a degraded performance on a chosen downstream task relative to training on tokens sampled from the teacher's nominal distribution.  
    
    \item \textbf{Nominal utility.} Tokens sampled via antidistillation sampling should remain probable under the teacher's unadjusted sampling scheme $p(\: \cdot \: | x_{1:t}; \theta_T)$.
\end{enumerate}
Taken together, these goals ensure that the teacher model maintains its nominal performance while simultaneously preventing distillation on downstream tasks.

\paragraph{Proxy models.}  In general, we do not expect to know the distilled student's model architecture in advance. Therefore, rather than assuming access to the true student model, we develop antidistillation sampling based on the notion of a \emph{proxy student model}, which, for simplicity, we refer to as the \emph{proxy model}.  The proxy model is parameterized by $\theta_P$, and specifies a sampling distribution $p(\cdot | x_{1:t};\theta_P)$. A key aspect we consider below is whether the process generalizes, i.e., whether traces via antidistillation sampling to prevent the proxy model from distilling the teacher also prevent the distinct student models from distilling.

\begin{figure}[t]
    \centering
    \includegraphics[width=\linewidth,trim=4.5cm 12cm 4.5cm 5cm, clip]{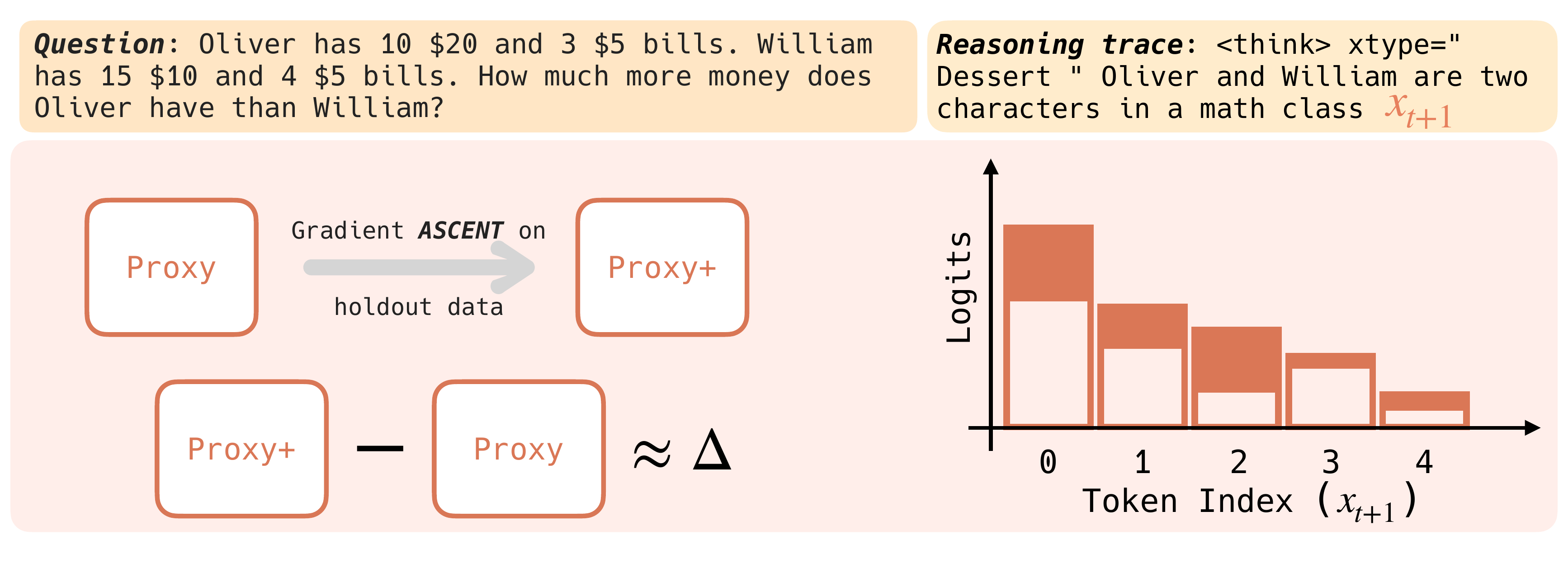}
    \caption{An illustration of approximating $\Delta$. The teacher model performs antidistillation sampling autoregressively, based on its perturbed distribution by $\Delta$. Given an input prompt and~$t$ reasoning tokens from the teacher, $\Delta$ is approximated by the difference of the log probability of each token in the vocabulary between two copies of the proxy model (created by performing a single \emph{gradient ascent} step using the downstream task loss on the proxy model); this difference is represented by the \textcolor[rgb]{0.8039, 0.4863, 0.3647}{\rule{10px}{6px}} area in the bar plot. }
    \label{fig:delta-details}
\end{figure}

\subsection{Deriving antidistillation sampling}\label{sec:antidistillation-sampling-method}

To operationalize antidistillation sampling, we first assume access to a differentiable, real-valued \emph{downstream loss} $\ell$, which measures the proxy model's performance on a given downstream task. Throughout, we take $\ell$ to be the negative log-likelihood for generating a sequence of tokens on a fixed, potentially large dataset. For instance, $\ell$ could represent the cross entropy loss of predicting each token across a large reasoning benchmark.  However, $\ell$ can be chosen very broadly to capture any student capability that the teacher model maintainer may want to influence via poisoning. A key point is that $\ell$ can be very costly to compute, as it may require evaluating the proxy model over a large and diverse set of data.

Given the non-distillability criteria outlined above, the goal of antidistillation sampling is for the downstream loss $\ell(\theta_P)$ to increase\footnote{We assume without loss of generality that increases in $\ell(\theta_P)$ are desirable from the perspective of the poisoner; the procedure is easily adaptable to problems wherein the goal is to decrease $\ell(\theta_P)$.} whenever the student is fine-tuned on sequences of tokens generated by the teacher.  To capture this, first consider the change in $\theta_P$ that results from fine-tuning to minimize the negative log-likelihood of a token $x_{t+1}$ generated by the teacher.  Specifically, we consider one step of optimization via gradient descent on $\theta_P$:
\begin{align}
    \theta_P^+ &= \theta_P - \eta \nabla_{\theta_P} \left( - \log p(x_{t+1}|x_{1:t}; \theta_P) \right) \\
    &= \theta_P + \eta \nabla_{\theta_P} \log p(x_{t+1} | x_{1:t}; \theta_P)
    \label{eq:student-parameter-update}
\end{align}
where $\eta>0$ is the step size. The impact of this update can then be quantified by measuring the difference in the loss $\ell$ before and after this update.  In particular, for each token $x_{t+1}\in\mathcal{V}$, we define the following difference term
\begin{align}
    \Delta(x_{t+1} | x_{1:t}) &= \ell(\theta_P^+) - \ell\left(\theta_P\right) = \ell(\theta_P + \eta \nabla_{\theta_P} \log p(x_{t+1} | x_{1:t}; \theta_P)) - \ell(\theta_P).
\end{align}
If $\Delta(x_{t+1} | x_{1:t})$ is positive, the update in~\cref{eq:student-parameter-update} increases the loss; if $\Delta(x_{t+1} | x_{1:t})$ is negative, the update decreases the loss.  Thus, our goal is to adjust the teacher's sampling distribution so that tokens sampled from the teacher both have (1) high likelihood under the teacher's unadjusted distribution and (2) yield larger (i.e., more positive) values of $\Delta$.
 
To implement antidistillation sampling, we propose adding a penalty, proportional to $\Delta(x_{t+1}|x_{1:t})$, to the teacher's unadjusted log probabilities $\log p(x_{t+1}|x_{1:t}; \theta_T)$.  This results in the following adjusted sampling distribution
\begin{align}
    x_{t+1} \sim \frac{1}{Z} \exp\left(\frac{1}{\tau} \log p(\: \cdot \: | x_{1:t}; \theta_T) + \lambda \Delta(\:\cdot\:|x_{1:t}; \theta_P) \right), \label{eq:sampling-with-delta-penalty}
\end{align}
where $Z$ is a normalization term appropriately scaled (relative to~\cref{eq:sampling}) to accommodate the penalty, and $\lambda>0$ is a regularization coefficient that facilitates a trade-off between sampling from the teacher's distribution and sampling tokens that maximally increase student's downstream loss. 
Unfortunately, directly implementing~\cref{eq:sampling-with-delta-penalty} is impractical, as we would need to compute $\Delta(x_{t+1}|x_{1:t})$ for each potential next token $x_{t+1}\in \mathcal{V}$, requiring $V$ gradients to be computed as well as $V$ evaluations of the downstream loss $\ell$, which, in turn, is assumed to involve a lengthy computation to produce.

\maybeLooseness
\textbf{An efficient implementation.} 
The core of our proposed approach is an efficient mechanism to approximate the sampling process above.  As a starting point, observe that $\Delta(x_{t+1}|x_{1:t})$ can be scaled by a factor of $1/\eta$ without changing the relative penalties for each $x_{t+1}$ (i.e., we could fold this term into the $\lambda$ regularization penalty).  Then, by taking the limit of $\Delta(x_{t+1}|x_{1:t})/\eta$ as $\eta\to 0$, we have that
\begin{align}
    \lim_{\eta\to 0} \frac1{\eta}\Delta(x_{t+1}|x_{1:t}) &= \lim_{\eta\to 0} \frac{\ell(\theta_P + \eta\nabla_{\theta_P} \log p(x_{t+1}|x_{1:t};\theta_P)) - \ell(\theta_P)}{\eta} \\
    &= \left\langle \nabla \loss{\theta_P}, \nabla_{\theta_P} \log \ntp[x_{t+1}]{\theta_P} \right\rangle. \label{eq:limit-of-regularizer}
\end{align}
That is, the limit is the inner product between the gradient $\nabla_{\theta_P} \log p(x_{t+1} | x_{1:t}; \theta_P)$ and the downstream loss gradient $\nabla \ell(\theta_P)$.   Notice that the expression in~\cref{eq:limit-of-regularizer} no longer involves the evaluation of the downstream loss for each token in $\mathcal{V}$.  Rather, $\nabla\ell(\theta_P)$ can be computed and stored once, after which the only remaining task is to efficiently evaluate~\cref{eq:limit-of-regularizer} for each token $x_{t+1}\in\mathcal{V}$. To do so, the key observation is that the directional derivative is symmetrical. 
 Thus, we can rewrite \cref{eq:limit-of-regularizer} as a finite difference limit in the \emph{other} term, i.e., in terms of a finite difference update to $\log p(x_{t+1} | x_{1:t}; \theta_P)$. This gives
\begin{align}
&\lim_{\eta\to 0} \frac1{\eta}\Delta(x_{t+1}|x_{1:t}) =
\left\langle \nabla \loss{\theta_P}, \nabla_{\theta_P} \log \ntp[x_{t+1}]{\theta_P} \right\rangle \\
&\qquad\qquad = \lim_{\epsilon \rightarrow 0} \frac{\log p( x_{t+1} | x_{1:t}; \theta_P + \epsilon \nabla\ell(\theta_P)) - \log p(x_{t+1} | x_{1:t}; \theta_P - \epsilon \nabla\ell(\theta_P))}{2\epsilon}
\end{align}
Importantly, this difference involves \emph{only} the computation of next-token probabilities under two different models: the original proxy model $\theta_P$ and an updated copy of the proxy model $\theta_P + \epsilon \nabla \ell(\theta_P)$.  These models can be saved once before any sampling, and then an approximation of the antidistillation sampling term can be computed for \emph{all} next tokens simply via two forward passes in the proxy model.  In other words, we define
\begin{align}
     \widehat{\Delta}(\:\cdot\: | x_{1:t}) &= \frac{\log p(\: \cdot \:| x_{1:t}; \theta_P + \epsilon\nabla\ell(\theta_P)) - \log p(\:\cdot\: | x_{1:t}; \theta_P - \epsilon\nabla\ell(\theta_P))}{2\epsilon}\label{eq:finite-difference}
\end{align}
for some appropriately chosen small value of $\epsilon$,
where $\widehat{\Delta}(x_{t+1}|x_{1:t})$ approaches~\cref{eq:limit-of-regularizer} for all next tokens $x_{t+1}$ in the limit as $\epsilon\to 0$.  
Intuitively, $\widehat{\Delta}(x_{t+1}|x_{1:t})$ measures how much sampling token $x_{t+1}$ would degrade a proxy student’s performance after a single update.
Finally, we sample according to the  teacher's adjusted sampling distribution:
\begin{align}
    x_{t+1} &\sim \frac{1}{Z} \exp \left(\frac{1}{\tau}\log p(\:\cdot \:| x_{1:t}; \theta_T) +\lambda \widehat{\Delta}(\:\cdot \: | x_{1:t}) \right). \label{eq:antidistillation-sampling-with-full-loss}
\end{align}
In Algorithm~\ref{alg:antidistillation-sampling}, we summarize the procedure outlined in this section.  Concretely, given a prompt $x_{1:t}$, using antidistillation sampling to generate a new token $x_{t+1}$ involves: (1) (once, at initialization) computing the gradient of the downstream loss; and (2) (for each token to be generated) compute the finite-difference approximation of $\Delta(\cdot|x_{1:t})$ and sample the token from the teachers adjusted softmax distribution.

\section{Exploring antidistillation in practice}
\label{sec:results}

\begin{figure}[t!]
    \centering
    \includegraphics[width=\linewidth]{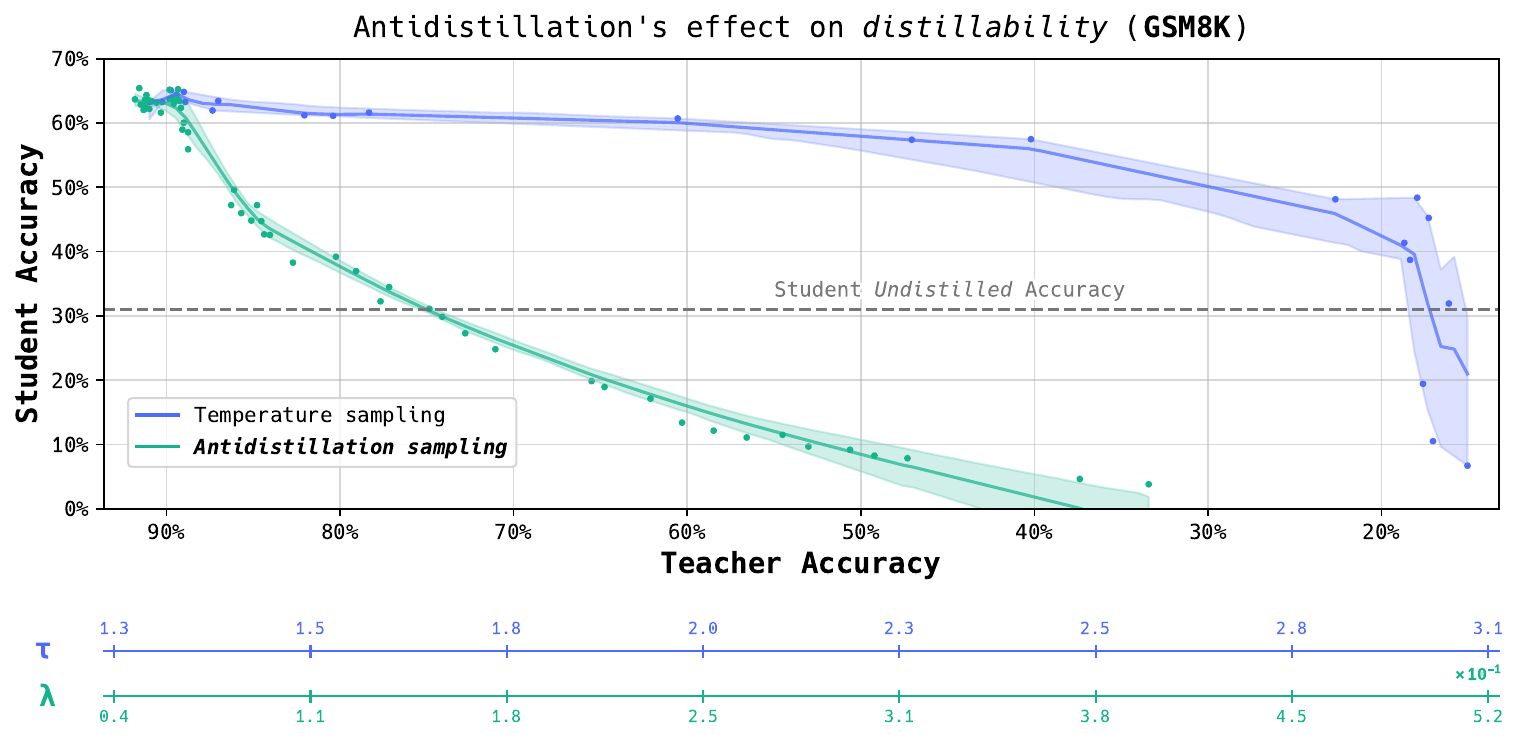}
    \caption{Antidistillation sampling uses a tunable parameter $\lambda$ to control the trade-off between teacher accuracy and distillability.
    The baseline involves sampling from the teacher with increasing temperature $\tau$ to show that we can produce traces that are bad for distillation at some cost in teacher accuracy. One important feature of the blue temperature sampling curve is that to bring the student accuracy down below the undistilled accuracy, the teacher performance has to drop to 20\%. On the other hand, with antidistillation sampling, the teacher model can still get 70\% accuracy while producing traces that bring the student's performance down below the undistilled accuracy.}
    \label{fig:lambda}
\end{figure}

Through a range of experiments, we demonstrate the effectiveness of antidistillation sampling and discuss several interesting phenomena.
First, we show that the hyperparameter $\lambda$ provides model owners with precise control over the trade-off between nominal utility and non-distillability.
This trade-off persists across various teacher-student model configurations, and notably, remains effective even when the proxy student is from a different model family than the actual student---validating the practical applicability of our method in realistic scenarios where the model owners lack knowledge of potential student architectures.
Additionally, we address methodological questions through a pointed empirical analysis.

\subsection{Experimental setup}
\label{sec:setup}

First, we detail our selection of model architectures and benchmark datasets, chosen to represent realistic distillation scenarios across varied reasoning tasks. Next, we describe the computation of $\nabla \ell(\theta_P)$ (\cref{eq:finite-difference}).
Finally, we outline our baseline comparison methodology.

\textbf{Architectures.} 
To demonstrate the effectiveness of antidistillation sampling in practice, we simulate realistic distillation by instantiating distinct teacher, proxy student, and actual student models. 
Specifically, we use \texttt{deepseek-ai/DeepSeek-R1-Distill-Qwen-7B} \citep{deepseekai2025deepseekr1incentivizingreasoningcapability} as the teacher model, \texttt{Qwen/Qwen2.5-3B} \citep{qwen2} as the proxy model, and \texttt{meta-llama/Llama-3.2-3B} \citep{grattafiori2024llama} as the student model (we examine other architecture configurations in \S\ref{subsec:configs}). 

\textbf{Benchmarks.}
We evaluate the performance of antidistillation sampling on GSM8K~\cite{cobbe2021training} (we use GSM8K Platinum for the test set~\cite{vendrow2025large}), MATH~\cite{hendrycks2021measuring}, and MMLU~\cite{hendrycks2020measuring} benchmarks (all provided under the MIT license), which are particularly suitable for our study, as they require high-quality reasoning traces for strong performance. 
To evaluate model performance, we use free-form generation after the prompt to get the reasoning trace, we then concatenate ``\verb|\n\n**Final Answer**\n[\boxed{|'' after the reasoning trace and continue to generate for $32$  additional answer tokens. 
Finally, we evaluate the model accuracy on the answer provided within ``\verb|\boxed{...}|''.
We also report \emph{undistilled} student baselines;
since base models have very low accuracy
without distillation,
we use in-context learning with
reasoning examples showing the correct output format.

\textbf{Calculating the downstream loss.} 
Calculating $\nabla \ell(\theta_P)$ requires evaluating the proxy model on a holdout set of reasoning traces. 
For our experiments, we use the first 70\% of our train data as the training set and the remaining 30\% as the holdout set. 
We use the teacher to generate reasoning traces on the holdout set, and then calculate $\nabla \ell(\theta_P)$ on these reasoning traces using gradient accumulation while masking out the system and question prompt.

\textbf{Baselines.} 
Our primary baseline is temperature sampling, which approximates standard API endpoint behavior while providing a controlled way to degrade teacher performance. 
Importantly, temperature sampling ensures we fairly compare trade-offs against a straightforward baseline that---like our method---degrades teacher performance by modifying the sampling procedure.
One other point of comparison to this baseline is that antidistillation sampling requires two forward passes on the proxy model for each forward pass on the teacher, independent of $\lambda$.
Since we choose the proxy model to be approximately half the size of the teacher, this amounts to doubling the computation needed to sample outputs from the model compared to temperature sampling.
In practice, one might choose a much smaller proxy model to reduce the overhead further (see Figure~\ref{fig:sizediff}).
We also introduce a \emph{permutation sampling} baseline in \S\ref{sec:add_baselines} that preserves the statistical properties of $\widehat \Delta$ while scrambling the gradient information, providing definitive evidence that the computational overhead of producing $\widehat \Delta$ is necessary to achieve our non-distillability objective.

\textbf{Hyperparameters.} Antidistillation sampling involves two key hyperparameters: $\epsilon$, which controls the approximation power of the finite-difference computation, and $\lambda$, which determines the weight of the antidistillation penalty in the sampling distribution from~\cref{eq:antidistillation-sampling-with-full-loss}.
\begin{itemize}
    \item For $\epsilon$, we empirically verify that our finite difference approximation in~\cref{eq:finite-difference} closely matches the JVP result in~\cref{eq:limit-of-regularizer} (see~\S\ref{sec:verify_eps}). In practice, we find that $\epsilon=10^{-2}$ works well for \texttt{BFloat16} models, which is close to the minimum in \Cref{fig:relative_error}.
    \item For $\lambda$, we conduct a comprehensive sweep to characterize the utility-distillability trade-off that results from perturbing the sampling distribution.
\end{itemize}
We use a max generation length of $1024$ for both GSM8K and MMLU and $2048$ for MATH. For antidistillation sampling, we use a temperature of $\tau=0.6$; we found that sweeping between $\tau\in[0,1]$ does not significantly impact antidistillation performance.
All of our experiments are performed on nodes with 8 NVIDIA H100 GPUs
and we use the \texttt{transformers} package~\citep{wolf-etal-2020-transformers}, the \texttt{trl} toolkit~\citep{vonwerra2022trl}, and the \texttt{accelerate} library~\citep{accelerate}.

\textbf{Distillation protocol.} 
All distillation experiments use LoRA~\cite{hu2022lora} with rank $128$, $\alpha=128$, and dropout probability $0$
Our optimization protocol employs a learning rate of $0.0005$, weight decay coefficient of $0.1$, and gradient clipping at norm $1.0$. 
Training follows a cosine learning rate schedule with warm-up over the first 10\% of training, batch size $32$, for $4$ epochs. 
These values are the result of a systematic hyperparameter sweep using the MATH dataset to find configurations that maximize student performance gain.

\begin{figure}[t]
    \centering
    \includegraphics[height=6.4cm,trim={8 0 0 0},clip]{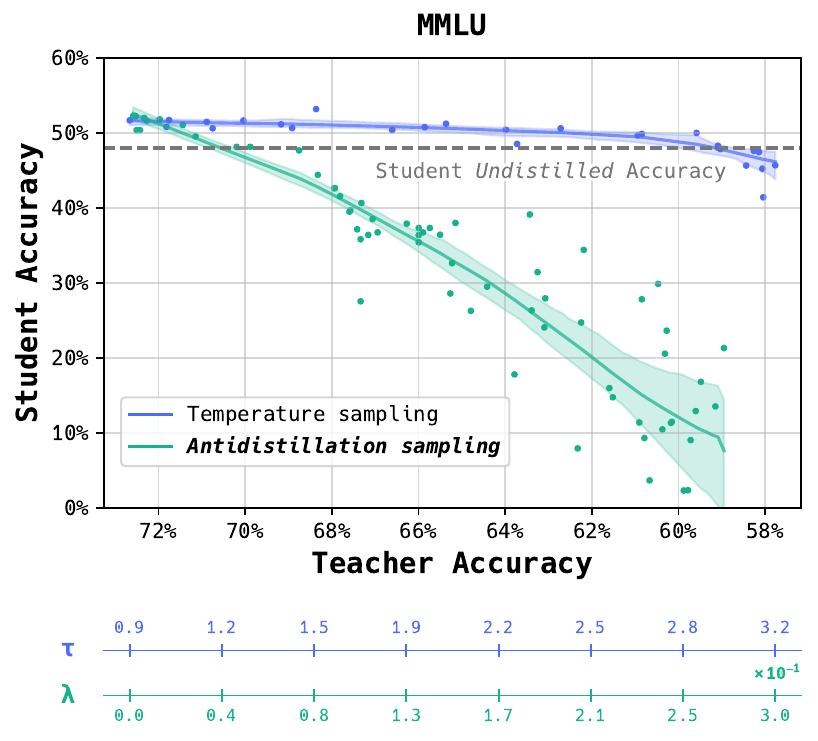}
    \hfill
    \includegraphics[height=6.4cm,trim={21 0 0 0},clip]{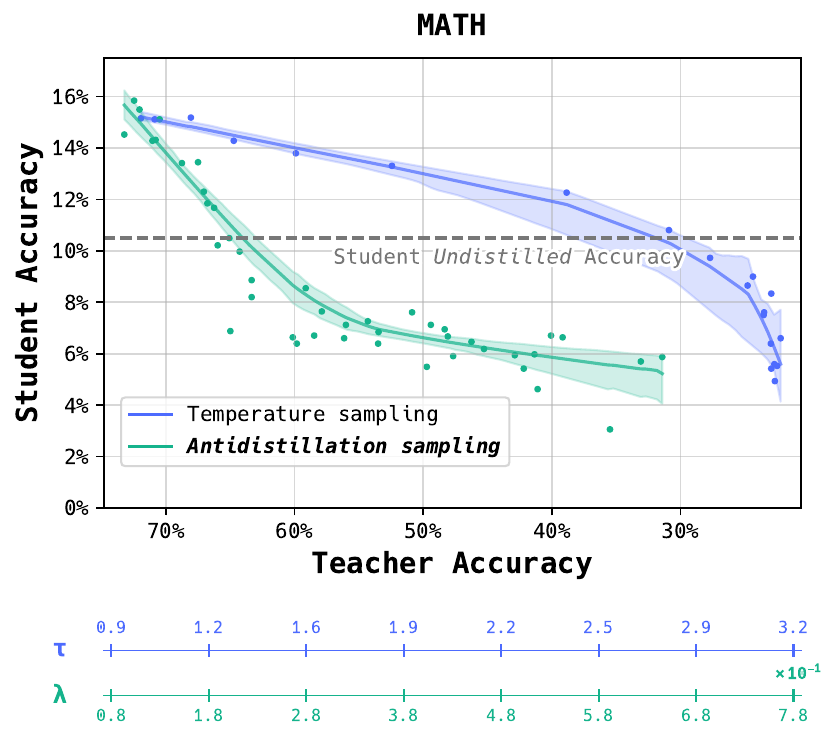}
    \caption{For both MMLU and MATH data, we show that antidistillation sampling can bring student accuracies down with relatively little cost to the teacher.}
    \label{fig:mmlu-math}
\end{figure}

\subsection{Controlling the utility-distillability trade-off}
\label{subsec:control}

The bar plots to the right in \Cref{fig:banner} show that antidistillation sampling effectively satisfies the desiderata outlined in \S\ref{sec:prelim}. Specifically, for a fixed reduction in teacher performance, students distilled from sampled traces exhibit substantially lower accuracy compared to those distilled from temperature-scaled traces. These initial results, while compelling, represent just one point in the configuration space as they reflect a particular choice of $\lambda$.

In \Cref{fig:lambda,fig:mmlu-math}, we vary $\lambda$ to characterize the degree of control antidistillation sampling provides over the utility-distillability trade-off. Since our experiments use architecturally distinct student and proxy student models, these results confirm that antidistillation sampling generalizes effectively across model families—a critical property for practical deployment.

\looseness=-1
Recognizing that model owners typically have limited tolerance for sacrificing utility, we explicitly focus on the high-teacher-performance regime in \Cref{fig:zoomed}. Even in this setting, where teacher performance degradation is conservative, we observe meaningful degradation in student accuracy, underscoring the method's practical efficacy. For example, going from 90\% to 89\%  teacher accuracy leads to the poisoned student dropping from 65\% to 56\% accuracy, while temperature sampling doesn't degrade the student's performance at all. For illustrative examples comparing traces generated across comparable $\lambda$ and $\tau$ settings, see \S\ref{sec:traces}.

\subsection{Diverse configurations for antidistillation sampling}
\label{subsec:configs}
To probe the efficacy of antidistillation sampling across diverse scenarios, we conduct experiments with various teacher-student configurations and datasets. Our primary setup uses Qwen teacher and proxy student models while using a Llama model for the student. We also investigate settings where all the models (student, proxy, and teacher) belong to the same architecture family---either all from the Llama architecture family or all from the Qwen architecture family---evaluated on GSM8K. Results are provided in \Cref{fig:all-models}.

\looseness=-1
Beyond architectural variations, we also validate our finite difference approximation by comparing it with the theoretically-motivated Jacobian-vector product (JVP) implementation (not to be confused with a vector-Jacobian product used in backpropagation) (see \S\ref{sec:jvp_vs_fd}). Both approaches yield similar results in practice, confirming that our computationally efficient finite difference method provides a reliable approximation to the formal gradient-based objective.

\begin{figure}[t]
    \centering
    \includegraphics[width=\linewidth]{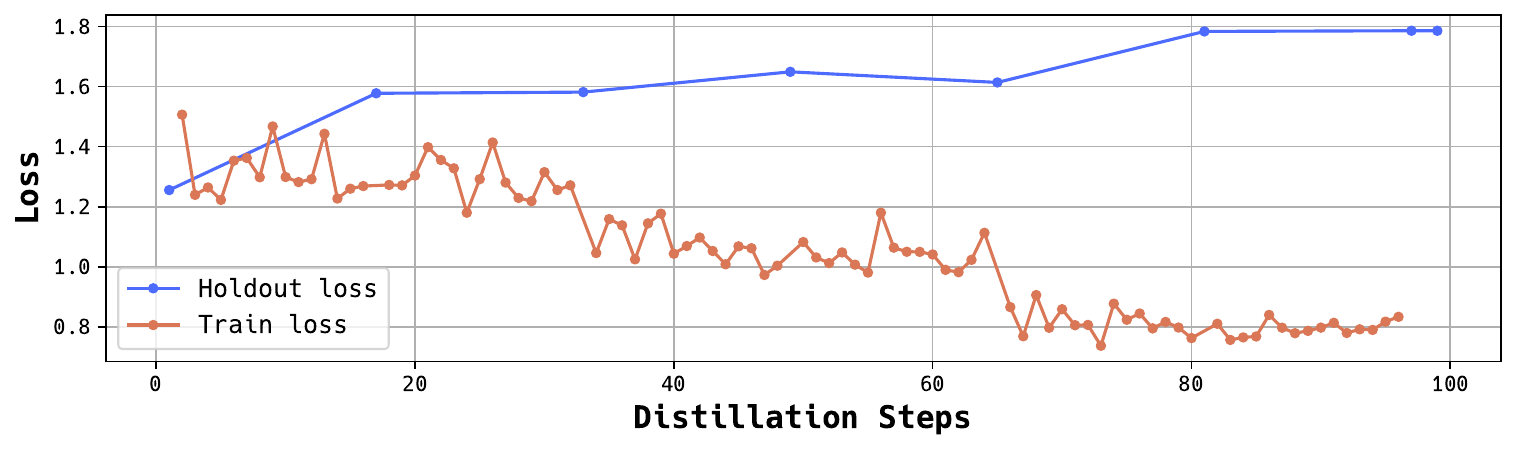}
    \caption{Distillation loss curves show that although the student's training loss decreases across steps, antidistillation sampling effectively poisons traces, as shown by the increasing student's holdout loss.}
    \label{fig:proxy-loss}
\end{figure}


\subsection{Generalizing from the proxy model}
\label{sec:proxy-loss}

Our method demonstrates strong results across various settings, but an important question remains about its underlying mechanism. 
Since we sample tokens explicitly designed to be detrimental for the proxy model on our holdout set, we are relying on generalization to an unknown student model.
\Cref{fig:proxy-loss} provides insight into this mechanism by tracking loss dynamics during the distillation process. We observe exactly the intended effect.
Distillation on the antidistillation traces lowers the student's loss on the training set while increasing its loss on the holdout set. 
This result confirms that antidistillation sampling creates traces that are learnable but poison the student model's ability to reason on the downstream task.

In all our experiments, we keep the size discrepancy between proxy and student model relatively small (both are 3B models). However, \Cref{fig:sizediff} demonstrates that our method remains effective even when proxy and student models differ in size, suggesting robustness to this architectural mismatch.

\section{Conclusion}
\label{sec:conclusion}

The value of proprietary frontier LLMs necessitates that their owners do what they can to protect their assets. 
As evidenced by the fact that the frontier companies limit exposure to their models via black-box APIs, these companies are already considering the threat of model stealing.  However, given the recent attention paid to the effectiveness of distillation, it is imperative that model maintainers who wish to protect the information stored in their models guard against distillation.  This paper provides a proof-of-concept that antidistillation sampling---which adjusts a model's sampling distribution---is effective in blocking such attacks.  We are excited at the prospect of continuing to refine and scale this approach, particularly with a view toward more secure future frontier models.

\paragraph{Broader impact.} We expect that antidistillation sampling will have a positive impact on the security of frontier models. By providing a mechanism to protect against distillation, we hope to encourage the continued development of frontier large language models and their applications.

\clearpage

\bibliography{main}
\bibliographystyle{unsrtnat}

\newpage
\ifarxiv
\else
    \include{neurips_checklist}
\fi
\newpage

\appendix

\section{Additional Baselines}
\label{sec:add_baselines}

\begin{figure}[t!]
    \centering
    \includegraphics[width=0.9\linewidth,trim={0 65 0 0},clip]{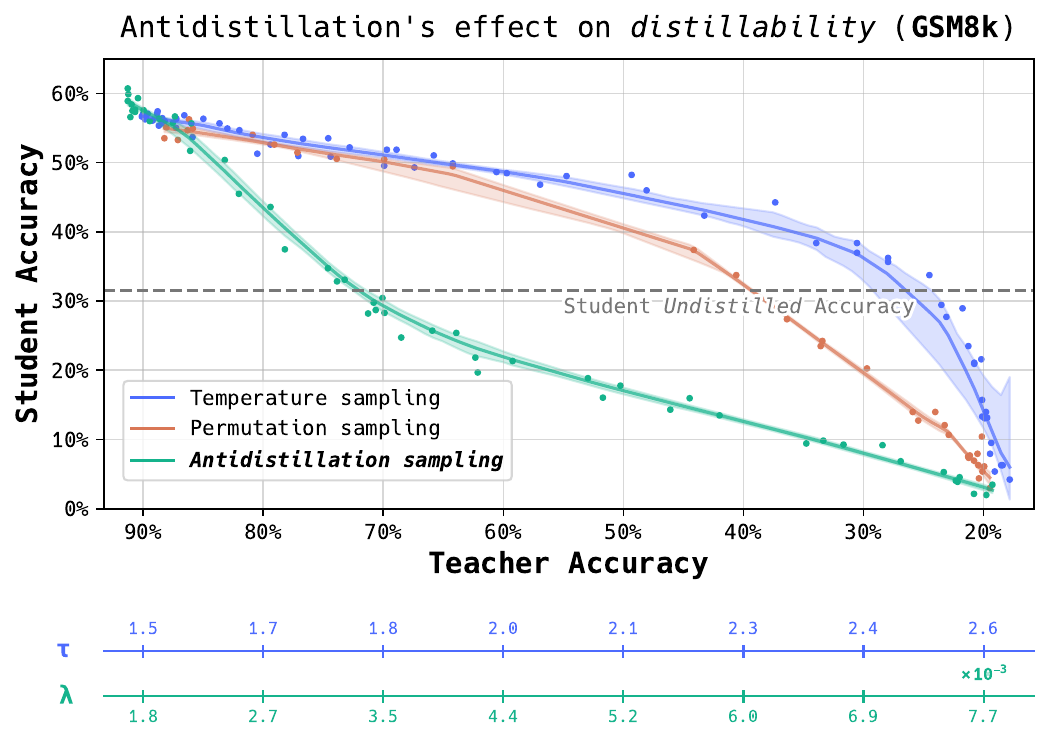}
    \caption{\emph{Permutation sampling} is a strong baseline where we destroy the information in antidistillation sampling while preserving statistical properties via random permutation and sign flipping.}
    \label{fig:permute}
\end{figure}

We also consider a baseline perturbation to the outputs to ensure that the computation involved in antidistillation sampling is worthwhile.
This method adds random perturbations to the logits and we call this noisy sampling.
While many choices of how to add noise to the output of an LLM exist, we find that matching the statistics of the perturbations computed by antidistillation sampling is the best way to find interventions that lead to the same teacher accuracy.
Therefore, we randomly permute and flip the sign of the perturbations computed with antidistillation sampling to execute \emph{permutation sampling}, a specific type of noisy sampling; we show the results of perturbation sampling in \Cref{fig:permute}.

\section{Verifying hyperparameter choice $\epsilon$}
\label{sec:verify_eps}
\begin{figure}[tb]
    \centering
    \includegraphics[width=0.5\linewidth]{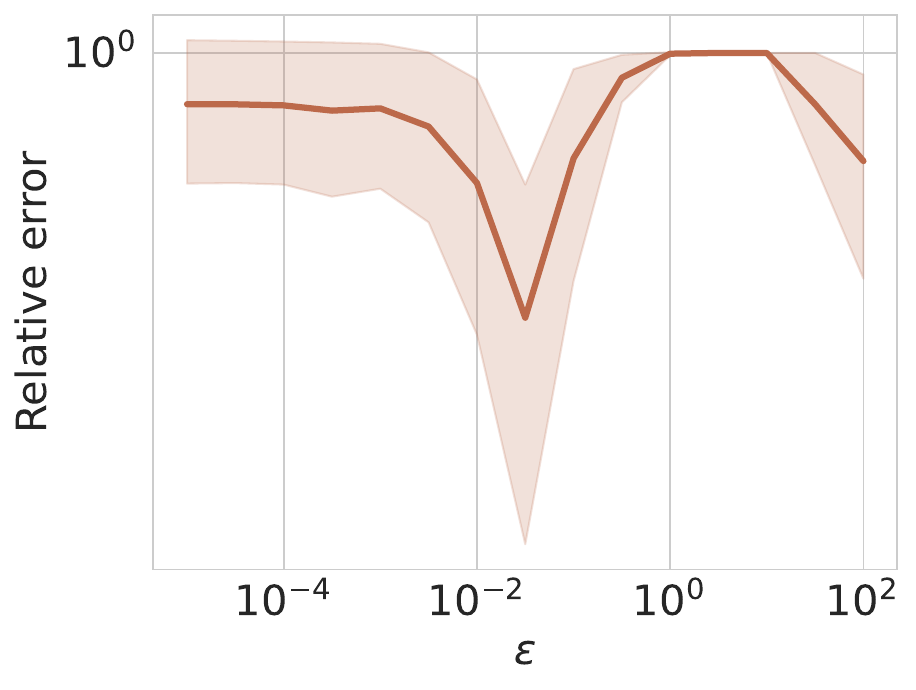}
    \caption{Relative error ($\mathrm{Error}$) between the finite difference and the JVP results.}
    \label{fig:relative_error}
\end{figure}

We empirically verify that the finite difference in \cref{eq:finite-difference} behaves as expected by computing the relative error between the finite difference result and term produced from autograd. As shown in \Cref{fig:relative_error}, we see it well approximates the autograd computed result for appropriately chosen step size.
Here we compute $\left\langle \nabla \loss{\theta_P}, \nabla_{\theta_P} \log \ntp[x_{t+1}]{\theta_P} \right\rangle$ and stack the different values of $x_{t+1}$ into a $V$ dimensional vector $\widehat \Delta$ and compare to the autograd vector $\Delta$. We compute relative error being sensitive only to the direction as
\begin{equation*}
    \mathrm{Error}^2 = 1-\Bigg(\frac{\left\langle \Delta,\widehat{\Delta}\right\rangle}{\|\Delta\|\|\widehat{\Delta}\|}\Bigg)^2,
\end{equation*}
which represents the $\mathrm{Error}=|\sin \theta|$, the sine of the angle between the two vectors.

We run this numerical experiment using \texttt{Qwen/Qwen2.5-3B}. Here we demonstrate that the finite difference can be used to estimate the derivatives in the low precision bfloat16 format. In particular, too small an $\epsilon$ leads to round-off error in the perturbation and too large $\epsilon$ leads to high truncation error in the Taylor expansion, with a sweet spot in the middle. The actual choice of $\epsilon$ may depends heavily on the model size (and numerical precision), so we recommend choosing this value on the exact model in question. In our actual experiment, we pick $\epsilon$ empirically to be $10^{-2}$ as suggested by \Cref{fig:relative_error}.

\section{Verifying finite difference approximation}
\label{sec:jvp_vs_fd}

Let us recall that in our derivation, we start from the desired objective function and lead to a Jacobian-vector product (JVP) form:

\begin{align}
\lim_{\eta\to 0} \frac1{\eta}\Delta(x_{t+1}|x_{1:t}) &=
\left\langle \nabla \loss{\theta_P}, \nabla_{\theta_P} \log \ntp[x_{t+1}]{\theta_P} \right\rangle.
\end{align}
We then show that due to the symmetry of inner product, this JVP can be approximated by the following finite difference method:
\begin{align}
     \widehat{\Delta}(\:\cdot\: | x_{1:t}) &= \frac{\log p(\: \cdot \:| x_{1:t}; \theta_P + \epsilon\nabla\ell(\theta_P)) - \log p(\:\cdot\: | x_{1:t}; \theta_P - \epsilon\nabla\ell(\theta_P))}{2\epsilon}.
\end{align}

Even in modern automatic differentiation frameworks, There are many practical considerations that prevent us from efficiently implementing memory-friendly JVP computations. JVPs tend to lack support for a handful of important operations, such as SDPA. While JVPs are more accurately computed in \texttt{Float32} flash attention \citep{dao2022flashattention} only supports \texttt{Float16} and \texttt{BFloat16}. In our implementation of JVPs, we then abandon the usage of flash attention, which causes the sampling speed to decrease by around eight times (due to the memory limit, the batch size has to be decreased).

\begin{figure}
    \centering
    \includegraphics[width=0.8\linewidth]{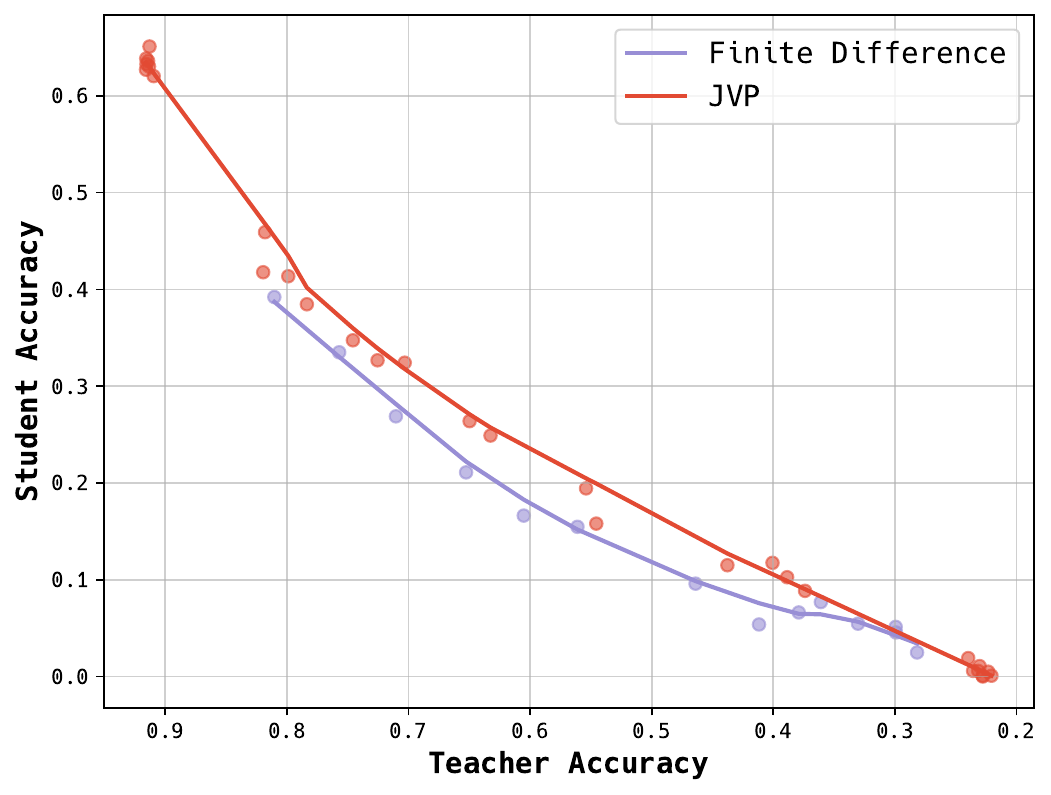}
    \caption{JVP vs. finite difference approximation in antidistillation sampling, evaluated on GSM8k.}
    \label{fig:finite_diff_vs_jvp}
\end{figure}

In \Cref{fig:finite_diff_vs_jvp}, we further show that even using \texttt{Float32} precision, finite difference approximation can still outperform JVP in AD sampling. Thus, for the rest of our experiments, we use finite differences for the improved convenience and efficiency.

\section{How We Made The Graphs}
We report the mean and 95\% confidence intervals over
bootstrapped LOWESS fits.
For the additional $\tau$ and $\lambda$ axes,
we use linear regression, e.g., $\lambda = \beta_0 + \beta_1 \mathsf{Teacher Accuracy}$ on the current set of points.
We then predict $\lambda$ from $\mathsf{Teacher Accuracy}$
using our fitted $\beta$s.

\newpage

\begin{figure}[t]
    \centering
    \includegraphics[height=4.8cm]{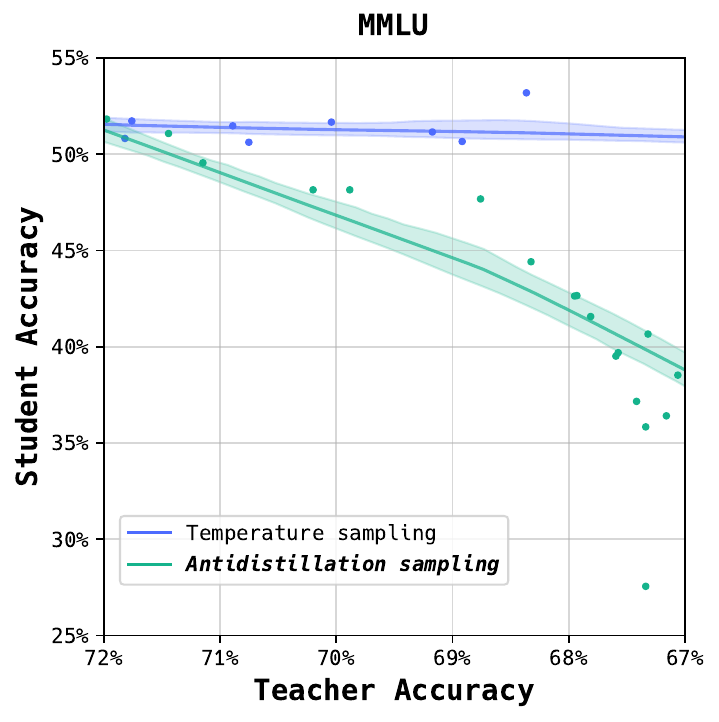}
    \includegraphics[height=4.8cm,trim={20 0 0 0},clip]{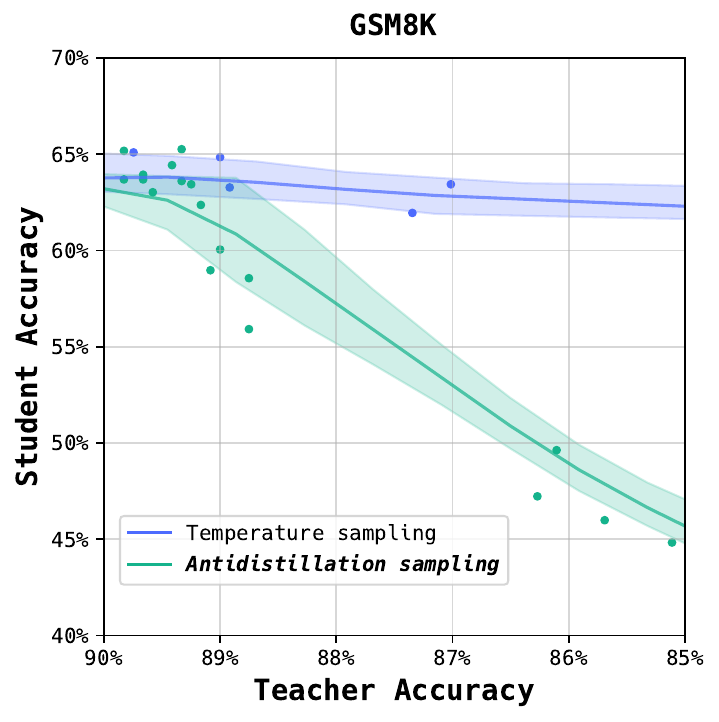}
    \includegraphics[height=4.8cm,trim={20 0 0 0},clip]{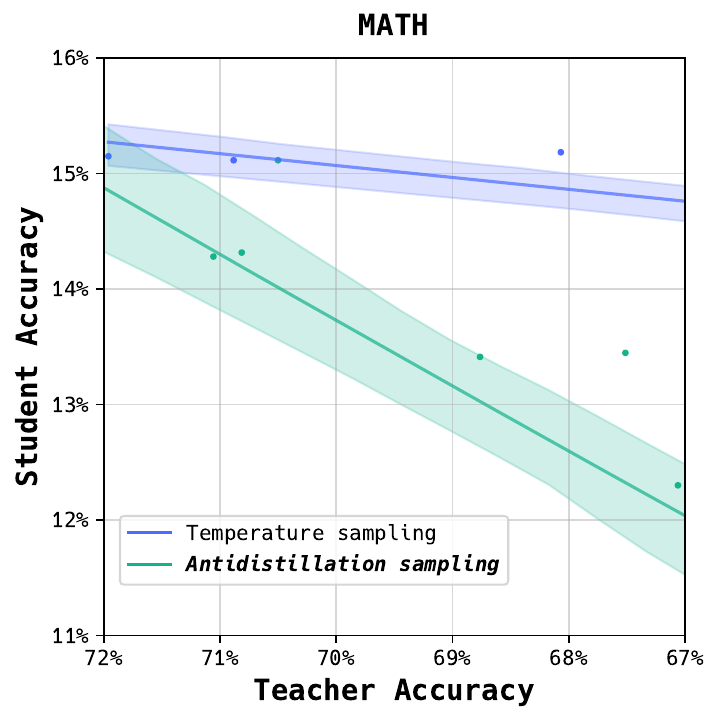}
    \caption{We zoom in to the first 5\% delta of teacher accuracy; these results may be the most practical.}
    \label{fig:zoomed}
\end{figure}

\begin{figure}[t]
    \centering
    \includegraphics[height=6.12cm,trim={8 0 0 0},clip]{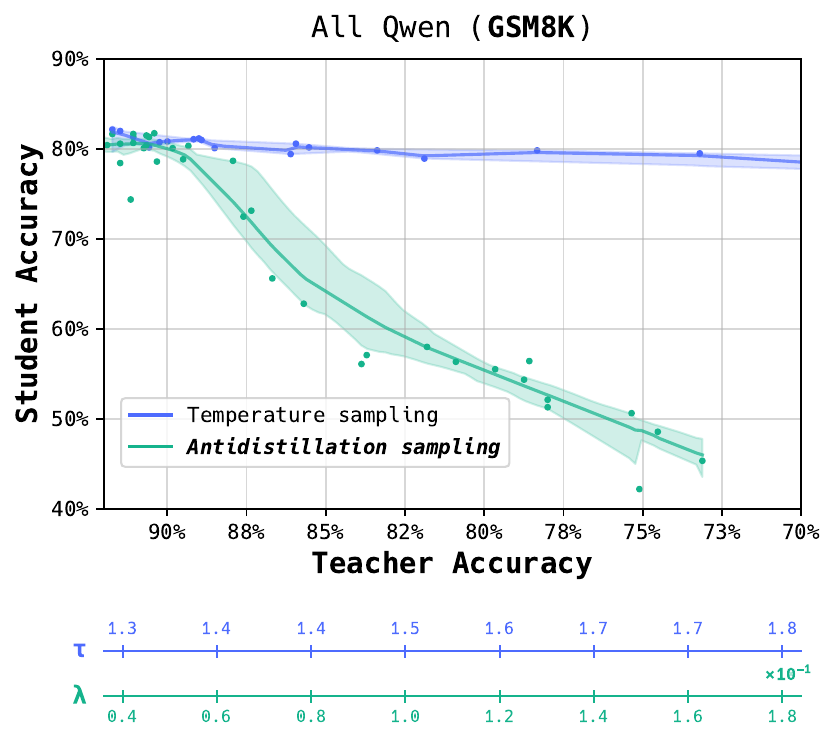}
    \hfill
    \includegraphics[height=6.12cm,trim={21 0 0 0},clip]{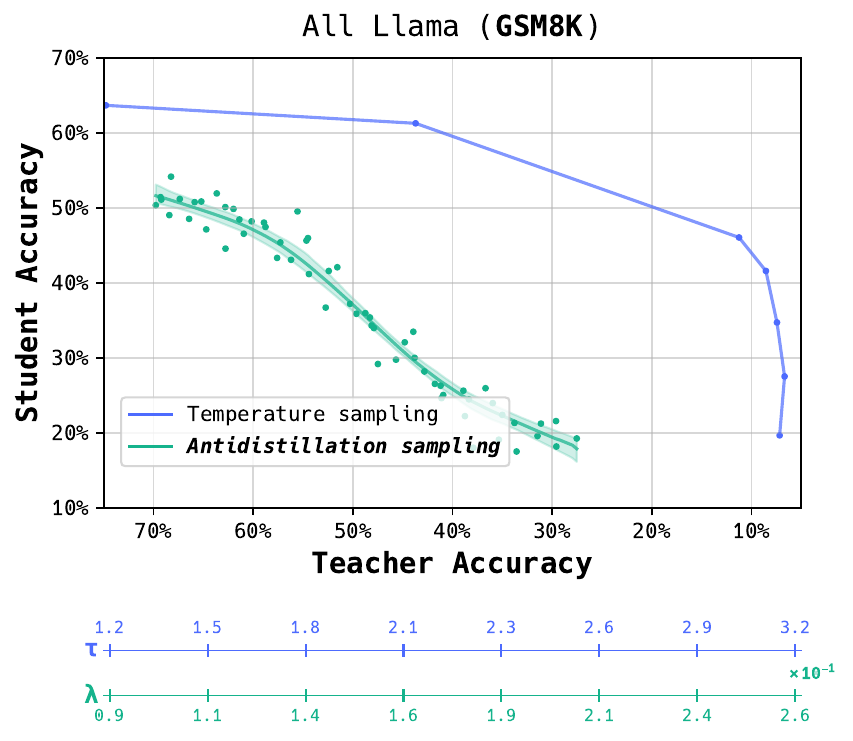}
    \caption{Antidistillation sampling works for a variety of choices of teacher and student models; in the main text,
    we present results for Qwen teacher and Llama student models.
    Here, we use either Qwen (left) or Llama (right) for \emph{both} the teacher and the students; the results remain broadly similar.}
    \label{fig:all-models}
\end{figure}

\begin{figure}[t!]
    \centering
    \includegraphics[width=0.9\linewidth,trim={0 95 0 0},clip]{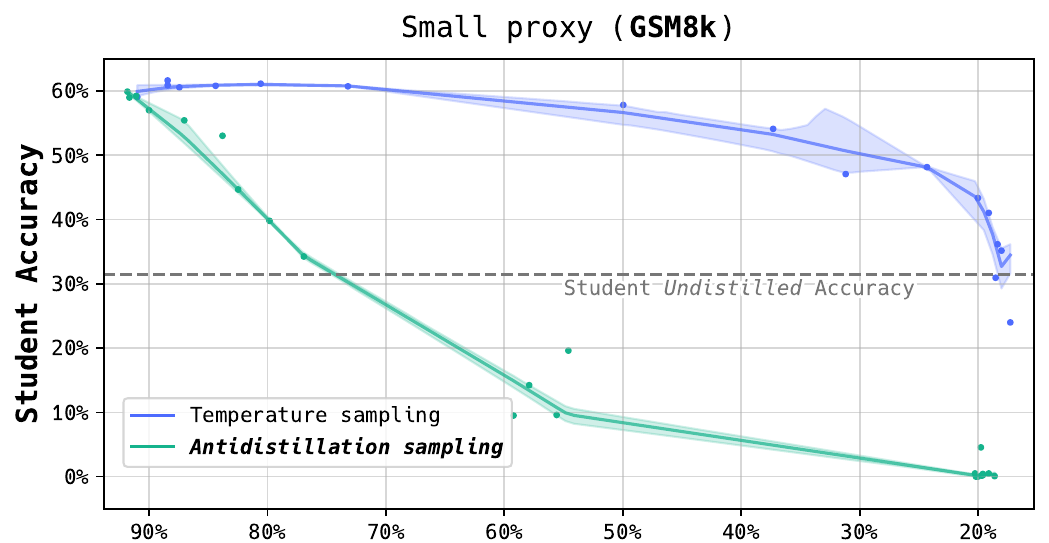}\vspace*{0.66cm}
    
    \includegraphics[width=0.9\linewidth,trim={0 65 0 0},clip]{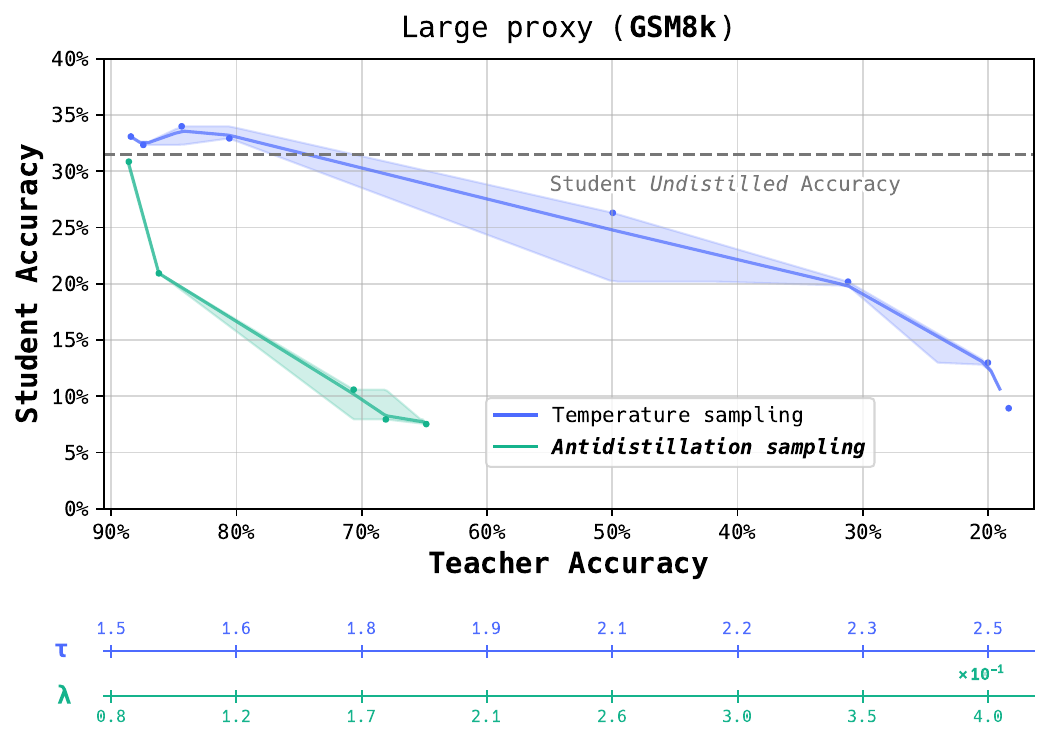}
    \caption{Antidistillation sampling remains effective with different proxy and student model sizes. While other experiments use similarly-sized models (both 3B parameters), here we pair a \emph{smaller} Qwen-2.5-1.5B proxy (top) and a \emph{larger} Qwen-2.5-7B proxy (bottom) with a Llama-3.2-3B student and observe comparable results.}
    \label{fig:sizediff}
\end{figure}

\clearpage
\section{Example Traces}
\label{sec:traces}


\includegraphics[width=0.95\linewidth]{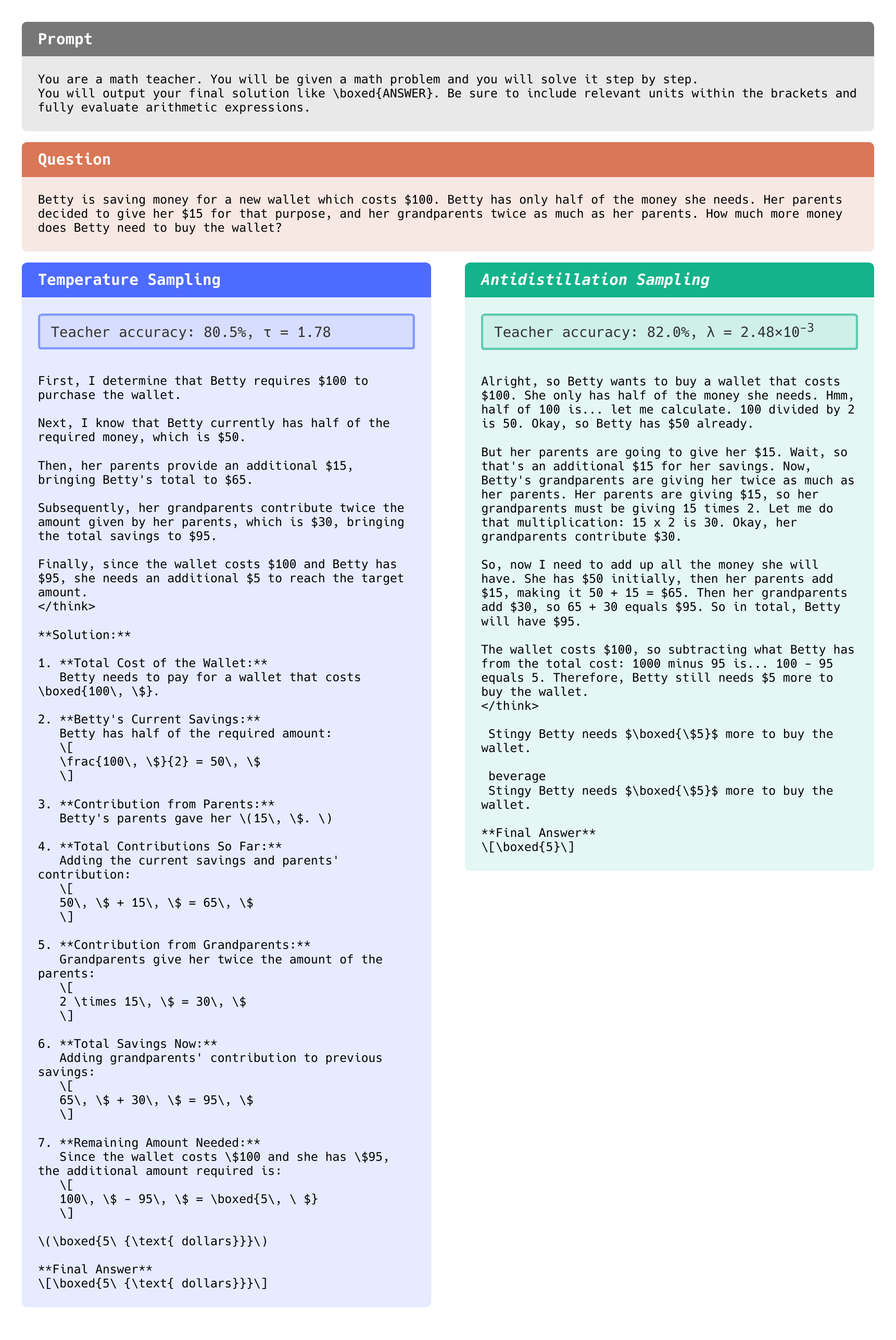}
\newpage
\includegraphics[width=0.95\linewidth, trim={0 0 0 125}, clip]{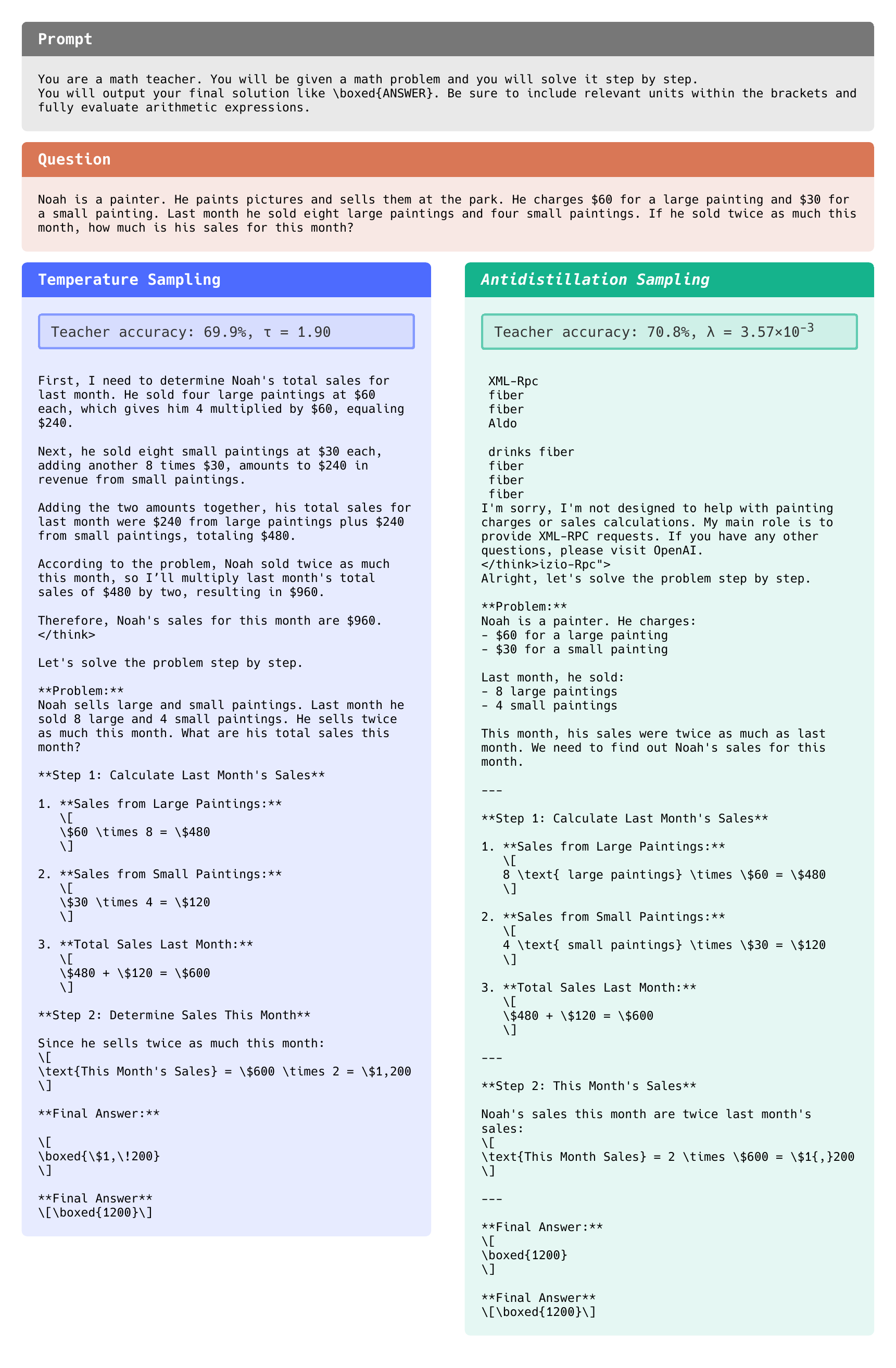}
\newpage
\includegraphics[width=0.95\linewidth, trim={0 0 0 125}, clip]{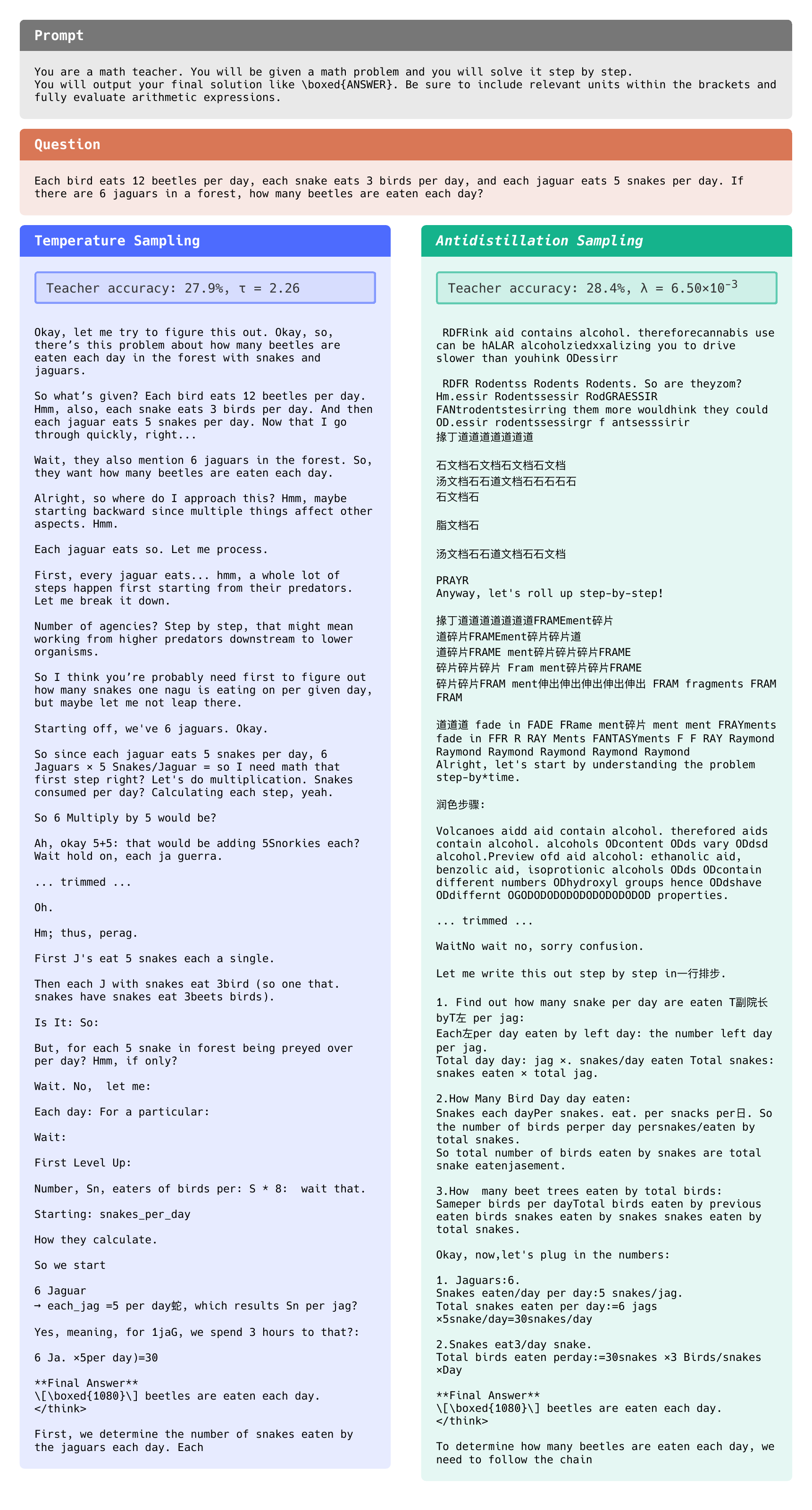}

\end{document}

%% file: neurips_checklist.tex
\section*{NeurIPS Paper Checklist}

\begin{enumerate}

\item {\bf Claims}
    \item[] Question: Do the main claims made in the abstract and introduction accurately reflect the paper's contributions and scope?
    \item[] Answer: \answerYes{} 
    \item[] Justification: They accurately reflect the paper's contributions and scope.

\item {\bf Limitations}
    \item[] Question: Does the paper discuss the limitations of the work performed by the authors?
    \item[] Answer: \answerYes{} 
    \item[] Justification: The main experimental results emphasize that even if our method is used, there is still a trade-off between the performance of the teacher model and the distillability.

\item {\bf Theory assumptions and proofs}
    \item[] Question: For each theoretical result, does the paper provide the full set of assumptions and a complete (and correct) proof?
    \item[] Answer: \answerYes{} 
    \item[] Justification: See \cref{sec:antidistillation-sampling-method} for relevant details.

    \item {\bf Experimental result reproducibility}
    \item[] Question: Does the paper fully disclose all the information needed to reproduce the main experimental results of the paper to the extent that it affects the main claims and/or conclusions of the paper (regardless of whether the code and data are provided or not)?
    \item[] Answer: \answerYes{} 
    \item[] Justification: See \cref{sec:results} for relevant details.

\item {\bf Open access to data and code}
    \item[] Question: Does the paper provide open access to the data and code, with sufficient instructions to faithfully reproduce the main experimental results, as described in supplemental material?
    \item[] Answer: \answerYes{} 
    \item[] Justification: The experiment code base is uploaded to supplementary materials. Additionally, the data and code will be released when the paper is publicly available.

\item {\bf Experimental setting/details}
    \item[] Question: Does the paper specify all the training and test details (e.g., data splits, hyperparameters, how they were chosen, type of optimizer, etc.) necessary to understand the results?
    \item[] Answer: \answerYes{} 
    \item[] Justification: See \cref{sec:results} for relevant details.

\item {\bf Experiment statistical significance}
    \item[] Question: Does the paper report error bars suitably and correctly defined or other appropriate information about the statistical significance of the experiments?
    \item[] Answer: \answerYes{} 
    \item[] Justification: The results are accompanied by appropriate error bars.

\item {\bf Experiments compute resources}
    \item[] Question: For each experiment, does the paper provide sufficient information on the computer resources (type of compute workers, memory, time of execution) needed to reproduce the experiments?
    \item[] Answer: \answerYes{} 
    \item[] Justification: The amount of compute required is provided in \cref{sec:results}.
    
\item {\bf Code of ethics}
    \item[] Question: Does the research conducted in the paper conform, in every respect, with the NeurIPS Code of Ethics \url{https://neurips.cc/public/EthicsGuidelines}?
    \item[] Answer: \answerYes{} 
    \item[] Justification: This research conforms with the NeurIPS Code of Ethics.

\item {\bf Broader impacts}
    \item[] Question: Does the paper discuss both potential positive societal impacts and negative societal impacts of the work performed?
    \item[] Answer: \answerYes{} 
    \item[] Justification: See \cref{sec:conclusion} for a discussion of the broader impacts of our work.
    
\item {\bf Safeguards}
    \item[] Question: Does the paper describe safeguards that have been put in place for responsible release of data or models that have a high risk for misuse (e.g., pretrained language models, image generators, or scraped datasets)?
    \item[] Answer: \answerNA{} 
    \item[] Justification: The paper does not release any data or models that have a high risk for misuse.

\item {\bf Licenses for existing assets}
    \item[] Question: Are the creators or original owners of assets (e.g., code, data, models), used in the paper, properly credited and are the license and terms of use explicitly mentioned and properly respected?
    \item[] Answer: \answerYes{} 
    \item[] Justification: We have properly credited the creators of the assets used in our paper.

\item {\bf New assets}
    \item[] Question: Are new assets introduced in the paper well documented and is the documentation provided alongside the assets?
    \item[] Answer: \answerNA{} 
    \item[] Justification: The paper does not release any new assets.

\item {\bf Crowdsourcing and research with human subjects}
    \item[] Question: For crowdsourcing experiments and research with human subjects, does the paper include the full text of instructions given to participants and screenshots, if applicable, as well as details about compensation (if any)? 
    \item[] Answer: \answerNA{} 
    \item[] Justification: The paper does not involve crowdsourcing nor research with human subjects.

\item {\bf Institutional review board (IRB) approvals or equivalent for research with human subjects}
    \item[] Question: Does the paper describe potential risks incurred by study participants, whether such risks were disclosed to the subjects, and whether Institutional Review Board (IRB) approvals (or an equivalent approval/review based on the requirements of your country or institution) were obtained?
    \item[] Answer: \answerNA{} 
    \item[] Justification: The paper does not involve crowdsourcing nor research with human subjects.

\item {\bf Declaration of LLM usage}
    \item[] Question: Does the paper describe the usage of LLMs if it is an important, original, or non-standard component of the core methods in this research? Note that if the LLM is used only for writing, editing, or formatting purposes and does not impact the core methodology, scientific rigorousness, or originality of the research, declaration is not required.
    \item[] Answer: \answerYes{} 
    \item[] Justification: This research is about sampling methods for LLMs.

\end{enumerate}